\pdfoutput=1

\documentclass[11pt]{article}

\usepackage{acl}

\usepackage{times}
\usepackage{latexsym}

\usepackage[T1]{fontenc}

\usepackage[utf8]{inputenc}

\usepackage{microtype}

\usepackage{inconsolata}

\usepackage{amsmath}
\usepackage{amssymb}
\usepackage{xspace}
\usepackage{url}
\usepackage{booktabs}
\usepackage{multirow}
\usepackage{multicol}
\usepackage{makecell}
\usepackage{xcolor}
\usepackage{graphicx}
\newcommand{\name}{CTC-S2UT\xspace}

\title{CTC-based Non-autoregressive Textless Speech-to-Speech Translation}

\author{
    Qingkai Fang$^{1,3}$,
    Zhengrui Ma$^{1,3}$,
    Yan Zhou$^{1,3}$,
    Min Zhang$^{4}$,
    Yang Feng$^{1,2,3}$\thanks{Corresponding author: Yang Feng.} \\
    \textsuperscript{\rm1}Key Laboratory of Intelligent Information Processing \\ Institute of Computing Technology, Chinese Academy of Sciences (ICT/CAS) \\
    \textsuperscript{\rm2}Key Laboratory of AI Safety, Chinese Academy of Sciences \\
    \textsuperscript{\rm3}University of Chinese Academy of Sciences, Beijing, China \\
    \textsuperscript{\rm4}School of Future Science and Engineering, Soochow University \\
    {$\;\:$\texttt{\{\href{mailto:fangqingkai21b@ict.ac.cn}{fangqingkai21b},\href{mailto:fengyang@ict.ac.cn}{fengyang}\}@ict.ac.cn}}
    {$\;\:$\texttt{\href{mailto:zhangminmt@hotmail.com}{zhangminmt}@hotmail.com}} \\
}

\begin{document}
\maketitle
\begin{abstract}
Direct speech-to-speech translation (S2ST) has achieved impressive translation quality, but it often faces the challenge of slow decoding due to the considerable length of speech sequences. Recently, some research has turned to non-autoregressive (NAR) models to expedite decoding, yet the translation quality typically lags behind autoregressive (AR) models significantly. In this paper, we investigate the performance of CTC-based NAR models in S2ST, as these models have shown impressive results in machine translation. Experimental results demonstrate that by combining pretraining, knowledge distillation, and advanced NAR training techniques such as glancing training and non-monotonic latent alignments, CTC-based NAR models achieve translation quality comparable to the AR model, while preserving up to 26.81$\times$ decoding speedup.\footnote{Code: \url{https://github.com/ictnlp/CTC-S2UT}.}

\end{abstract}

\section{Introduction}
Direct speech-to-speech translation (S2ST) refers to the process of generating target speech directly from the source speech, without the need for generating intermediate source or target text. Distinguished from traditional cascaded S2ST~\citep{599557, 1597243}, which involves cascading automatic speech recognition (ASR), machine translation (MT), and text-to-speech (TTS) models, direct S2ST can mitigate error accumulation and may be applicable to low-resource languages without written forms~\citep{hokkien}.

Early direct S2ST models often require text supervision during training~\citep{translatotron}, which is not feasible for numerous unwritten languages. To address this issue, researchers have proposed textless S2ST based on discrete units~\citep{s2ut, lee-etal-2022-textless}, which does not rely on any text supervision during training. Specifically, they first extract the discrete representation (i.e., discrete units) of the target speech based on the speech pretraining model HuBERT~\citep{hubert}. Then, they train a speech-to-unit translation (S2UT) model to generate target discrete units from the source speech and subsequently employ a unit-based vocoder to synthesize the target speech. However, due to the considerable length of discrete unit sequences, these models often exhibit higher decoding latency. To address this problem, \citet{huang2023chch} propose the use of non-autoregressive (NAR) model to generate discrete units in parallel. Despite its faster decoding speed, it suffers from a significant decrease in translation quality compared to S2UT limited by the expressive ability of NAR models.

In this paper, we aim to explore the performance of more powerful NAR models in S2ST. We opt for models based on connectionist temporal classification~\citep[CTC;][]{ctc}, which improves the model's expressive ability by expanding the output space. Previous work has demonstrated that CTC-based NAR models exhibit excellent performance in machine translation~\citep{libovicky-helcl-2018-end, yan-etal-2023-ctc} and speech-to-text translation~\citep{xu-etal-2023-ctc}. In this paper, we build upon CTC and incorporate multiple techniques including pretraining, knowledge distillation, and advanced NAR training techniques such as glancing training and non-monotonic latent alignments. Experimental results show that CTC-based models can attain translation quality comparable to the AR model, while maintaining up to 26.81$\times$ decoding speedup.

\section{Background}
\begin{figure*}[t]
    \centering
    \includegraphics[width=0.9\textwidth]{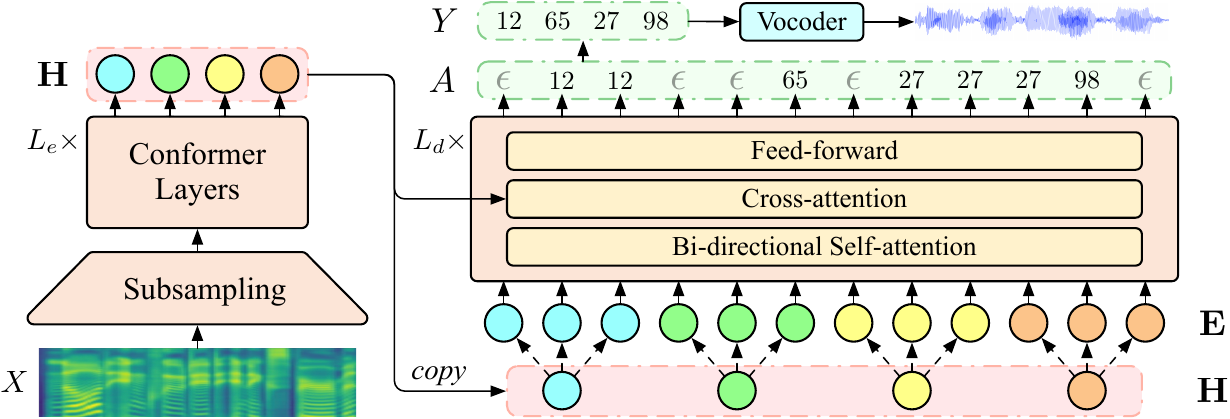}
    \caption{Model architecture of \name.}
    \label{fig:mode}
\end{figure*}

\subsection{Speech-to-unit Translation}
Speech-to-unit translation~\citep[S2UT;][]{s2ut} introduces \textit{discrete units} as training targets and has been proven to be one of the most effective approaches in textless S2ST. Specifically, discrete units are obtained by applying K-means clustering to the 50Hz continuous representation generated by the speech pretraining model HuBERT~\citep{hubert}. The cluster indices serve as the discrete units, ensuring that each discrete unit is an integer in the range of $0$ to $K-1$, where $K$ represents the number of clusters.

S2UT trains a sequence-to-sequence model to directly generate the discrete unit sequence corresponding to the target speech from the source speech. Finally, a pre-trained unit-based HiFi-GAN~\citep{hifi-gan} vocoder synthesizes the target speech waveform from the sequence of discrete units, completing the S2ST task.

\subsection{Connectionist Temporal Classification}
Connectionist temporal classification~\citep[CTC;][]{ctc} is a sequence modeling algorithm used to model \textit{variable-length} mappings between sequences, originally applied to speech recognition~\citep{pmlr-v32-graves14} and more recently extended to machine translation~\citep{libovicky-helcl-2018-end}. Formally, given an input sequence $X=(x_1, ..., x_N)$ and an output sequence $Y=(y_1, ..., y_M)$, CTC introduces a special blank token $\epsilon$ into the output space $\mathcal{Y}^*=\mathcal{Y}\cup\{\epsilon\}$. It first generates a sequence $A=(a_1, ..., a_T)$ based on the input sequence $X$, where $T$ is a pre-determined length, and $A$ is referred to as the \textit{alignment}. Furthermore, CTC defines a collapsing function $\beta(A)$, which first merges consecutive identical tokens in $A$ into one and then removes all blank tokens to obtain the final output sequence $Y$. During the training process, CTC marginalizes over all possible alignments for optimization:
\begin{equation}
    P(Y|X) = \sum_{A\in\beta^{-1}(Y)} P(A|X), 
    \label{eq:pyx}
\end{equation}
where $\beta^{-1}(Y)$ contains all possible alignments of length $T$ that can be collapsed to $Y$. The probability $P(A|X)$ is modeled in a non-autoregressive way:
\begin{equation}
    P(A|X) = \prod_{t=1}^T p(a_t | X).
\end{equation}

\section{Method}
In this section, we investigate the performance of CTC-based non-autoregressive translation models in S2ST. We develop our model based on S2UT, named \name, where $X$ represents the source speech, and $Y$ represents the sequence of discrete units corresponding to the target speech. Section \ref{sec:model} will introduce the model architecture, and Section \ref{sec:training} will introduce the training approach.

\subsection{Model Architecture}
\label{sec:model}

\name comprises a speech encoder and a non-autoregressive unit decoder. The speech encoder encodes the source speech, while the unit decoder generates the target discrete units in parallel.

\paragraph{Speech Encoder}
The speech encoder comprises a subsampling module and $L_e$ Conformer layers~\citep{conformer}. The subsampling module consists of two 1D convolutional layers, responsible for a $4\times$ subsampling of the input speech features. Conformer layers employ both attention mechanisms and convolutional layers to concurrently model global and local features of the speech. 
Relative positional encoding~\citep{dai-etal-2019-transformer} is employed in the multi-head attention. 
The output representations of the speech encoder are denoted as $\mathbf{H} = (\mathbf{h}_1, ..., \mathbf{h}_{N'})$, where $N'=\lfloor N/4 \rfloor$.

\paragraph{Unit Decoder}
The unit decoder consists of $L_d$ non-autoregressive Transformer decoder layers~\citep{gu2018nonautoregressive}, each comprising a bidirectional self-attention layer, a cross-attention layer, and a feed-forward layer. The input to the decoder is obtained by uniformly upsampling the encoder output representations by $\lambda$ times: $\mathbf{E} = (\mathbf{e}_1, ..., \mathbf{e}_T)$, where $T = \lambda\cdot N'$ and $\mathbf{e}_i = \mathbf{h}_{\lfloor i / \lambda \rfloor}$. We add learnable positional encodings to $\mathbf{E}$ and feed it into the decoder. The decoder generates the alignment $A$ in parallel, and finally obtains the sequence of target discrete units $Y=\beta(A)$.

\subsection{Training}
\label{sec:training}

In theory, we can calculate the conditional probability $P(Y|X)$ using Eq. (\ref{eq:pyx}) and train the model by minimizing $-\log P(Y|X)$. However, due to the complexity of the speech-to-unit translation task, training from scratch using a simple CTC objective often faces challenges. Therefore, we adopt several techniques to facilitate training, including encoder pretraining, knowledge distillation, glancing training and non-monotonic latent alignments.

\paragraph{Encoder Pretraining} We initialize the encoder of \name using the encoder from a pretrained autoregressive S2UT model to provide the encoder with a good initial state.

\paragraph{Knowledge Distillation} We perform sequence-level knowledge distillation~\citep[KD;][]{kim-rush-2016-sequence} with the autoregressive S2UT model, which is a widely used approach in non-autoregressive translation to reduce data multimodality~\citep{Zhou2020Understanding, ddrs}.

\paragraph{Glancing Training} Glancing training~\citep[GLAT;][]{qian-etal-2021-glancing} has been proven to significantly enhance the translation quality of NAR translation models. It first selects the alignment with the highest posterior probability: $A^* = \mathop{\arg\max}_{A\in \beta^{-1}(Y)} P(A|X)$, and then randomly chooses some positions in the decoder input $\mathbf{E}$ to replace with the corresponding token embeddings from the alignment $A^*$.

\paragraph{Non-monotonic Latent Alignments} \citet{nmla} introduces non-monotonic latent alignments (NMLA) to enhance CTC-based NAR models. It maximizes the F1 score of expected bigram matching between alignments and the target. More details can be found in the original paper.

\section{Experiments}
\subsection{Experimental Setups}
\begin{table}[t]
    \centering
\resizebox{\linewidth}{!}{
\begin{tabular}{c|c|c|c}
\toprule
\textbf{Direction} & \textbf{Source Hours} & \textbf{Target Hours} & \textbf{\#Samples} \\
\midrule
\textbf{Fr$\rightarrow$En} & 264h & 174h & 207K \\
\textbf{En$\rightarrow$Fr} & 174h & 264h & 207K \\
\textbf{En$\rightarrow$Es} & 70h & 113h  & 79K  \\
\bottomrule
\end{tabular}}
    \caption{Statistics of datasets. }
    \label{tab:data}
\end{table}

\paragraph{Dataset}
We use the CVSS-C dataset~\citep{jia2022cvss} for experiments, which includes speech-to-speech translation pairs between 21 languages and English. Following \citet{huang2023chch}, we conduct experiments on three language pairs, including French$\rightarrow$English (Fr$\rightarrow$En), English$\rightarrow$French (En$\rightarrow$Fr), and English$\rightarrow$Spanish (En$\rightarrow$Es). Table~\ref{tab:data} lists the detailed statistics of the dataset. We compute the 80-dimensional mel-filterbank features for the source speech, and extract the discrete unit sequences for the target speech using the publicly available pretrained mHuBERT model and K-means quantizer\footnote{\url{https://github.com/facebookresearch/fairseq/blob/main/examples/speech_to_speech/docs/textless_s2st_real_data.md}}.

\begin{table*}[t]
\centering
\small
\begin{tabular}{l|c|c|c|cccc|c}
\toprule
\multirow{2}{*}{\textbf{Model}} & \multirow{2}{*}{\textbf{\#Iter}} & \multirow{2}{*}{\textbf{\#Param}} & \multirow{2}{*}{\textbf{Textless}} & \multicolumn{4}{c|}{\textbf{ASR-BLEU}} & \multirow{2}{*}{\textbf{Speedup}} \\
& & & & \textbf{Fr$\rightarrow$En} & \textbf{En$\rightarrow$Fr} & \textbf{En$\rightarrow$Es} & \textbf{Avg.} &  \\
\midrule
 S2UT & $M$ & 58M & $\checkmark$ & 24.80  & 21.38 & 21.71 & 22.63 & 1.00$\times$ \\
\midrule
 TranSpeech & 5 & 67M & $\checkmark$ & 17.24 & 16.30 & 11.79 & 15.11 & 11.04$\times$ \\
 TranSpeech & 15 & 67M & $\checkmark$ & 18.03 & 16.97 & 12.62 & 15.87 & 5.34$\times$ \\
 TranSpeech (b=15) & 15 & 67M & $\checkmark$ & 18.10 & 17.05 & 12.70 & 15.95 & 2.75$\times$ \\
 TranSpeech (b=15 + NPD) & 15 & 67M & $\checkmark$ & 18.39 & 17.50 & 12.77 & 16.22 & 2.53$\times$ \\
\midrule
 DASpeech (Lookahead) & 1+1 & 93M & $\times$ & 24.71 & N/A & N/A & N/A & 18.53$\times$  \\
 DASpeech (Joint-Viterbi) & 1+1 & 93M & $\times$ & 25.03 & N/A & N/A & N/A & 16.29$\times$ \\
\midrule
 \name & 1 & 59M & $\checkmark$ & \textbf{25.16} & \textbf{20.85} & \textbf{21.60} & \textbf{22.54} & \textbf{26.81$\times$} \\
\bottomrule
\end{tabular}
\caption{Results on CVSS-C \texttt{test} sets. Results of TranSpeech~\citep{huang2023chch} and DASpeech~\citep{fang-etal-2023-daspeech} are quoted from the original paper. NPD: noisy parallel decoding; b: length beam in NAR decoding. The best results among NAR models are marked in \textbf{bold}.}
\label{tab:main-results}
\end{table*}

\paragraph{Model Configurations and Training}
For both autoregressive S2UT and \name, the models consist of 12 Conformer encoder layers and 6 Transformer decoder layers. The dropout is set to 0.3. We initialize the encoder of \name with the encoder from S2UT and train the model in two stages. In the first stage, we employ conventional CTC loss for training. Each batch contains 320k source audio frames, and the learning rate warms up to 1e-3 over 10k steps. The glancing ratio linearly decays from 0.5 to 0.3 within 100k steps. In the second stage, we fine-tune the model using NMLA loss for 6k steps. Each batch contains 1,280k source audio frames, and the learning rate warms up to 3e-4 over 500 steps. The glancing ratio is fixed at 0.3. The upsample factor $\lambda$ is set to 2, 6, and 6 for Fr$\rightarrow$En, En$\rightarrow$Fr, and En$\rightarrow$Es, respectively. We use Adam~\citep{adam} optimizer in all training stages. All models are trained on 4 RTX 3090 GPUs.

\paragraph{Evaluation and Baseline Systems}
We use the ASR-BLEU\footnote{\url{https://github.com/facebookresearch/fairseq/tree/ust/examples/speech_to_speech/asr_bleu}} toolkit to evaluate the translation quality. It transcribes the translated speech into text using a pretrained ASR model, and calculates the BLEU score~\citep{papineni-etal-2002-bleu}. We measure the decoding speedup on the CVSS-C Fr$\rightarrow$En \texttt{test} set using 1 RTX 3090 GPU with a batch size of 1. Besides autoregressive S2UT, we include TranSpeech~\citep{huang2023chch} and DASpeech~\citep{fang-etal-2023-daspeech} as NAR baseline systems for comparison.

\subsection{Main Results}

Table \ref{tab:main-results} presents the experimental results on the CVSS-C \texttt{test} set, where \name achieves translation quality comparable to autoregressive S2UT while preserving a significant decoding speedup of 26.81 times. Compared to the previous NAR S2UT model TranSpeech, which relies on CMLM~\citep{ghazvininejad-etal-2019-mask}, \name attains higher translation quality without the need for iterative decoding, demonstrating the importance of employing more powerful NAR models. Compared to the previous two-pass NAR model DASpeech, \name achieves slightly higher translation quality. Additionally, since \name does not require decoding the target text, it exhibits faster decoding speed and can be applied to target languages without written form.

\subsection{Ablation Study}
\begin{table}[]
    \centering
    \small
\begin{tabular}{cccc|c}
\toprule
\textbf{Pretrain} & \textbf{GLAT} & \textbf{KD} & \textbf{NMLA} & \textbf{ASR-BLEU} \\
\midrule
$\times$ & $\checkmark$ & $\checkmark$ & $\times$ & 0.00 \\
$\checkmark$ & $\times$ & $\checkmark$ & $\times$ & 11.44 \\
$\checkmark$ & $\checkmark$ & $\times$ & $\times$ & 23.19 \\
$\checkmark$ & $\checkmark$ & $\checkmark$ & $\times$ & 23.85 \\
$\checkmark$ & $\checkmark$ & $\checkmark$ & $\checkmark$ & \textbf{25.16} \\
\bottomrule
\end{tabular}
\caption{Results on CVSS-C Fr$\rightarrow$En \texttt{test} set with different combinations of training techniques.}
\label{tab:ablation}
\end{table}

Table \ref{tab:ablation} shows the results with different combinations of training techniques. We first explore the influence of encoder pretraining, GLAT and KD on training. The results indicate that encoder pretraining is indispensable, as the model fails to converge without it. GLAT also significantly impacts performance, while KD slightly enhances model performance. Building upon these techniques, finetuning the model using NMLA further improves performance, confirming the effectiveness of four training techniques we employed.

\subsection{Speedup Analysis}

We further investigate the decoding speedup ratio of \name compared to autoregressive S2UT for different lengths of speech. As shown in Table \ref{tab:speedup}, for longer source speech, the speedup ratio of \name becomes more significant, reaching up to 34.28 times, indicating the speed advantage of NAR models in translating long speech segments.

\begin{table}[]
    \centering
    \small
\begin{tabular}{c|ccc}
\toprule
\textbf{\#Frames} & $[0, 300)$ & $[300, 600)$ & $[600, +\infty)$ \\
\midrule
\textbf{Speedup} & 11.67$\times$ & 24.95$\times$ & \textbf{34.28$\times$} \\
\bottomrule
\end{tabular}
\caption{Speedup of \name compared with S2UT across different source speech lengths (\#Frames) on CVSS-C Fr$\rightarrow$ En \texttt{test} set.}
\label{tab:speedup}
\end{table}

\section{Related Work}
Speech-to-speech translation extends speech-to-text translation (S2TT)~\citep{fang-etal-2022-STEMM, fang-and-feng-2023-back, fang-and-feng-2023-understanding, zhou-etal-2023-cmot} that further synthesizes the target speech. \citet{translatotron} first introduces direct S2ST. \citet{s2ut, lee-etal-2022-textless, hokkien} use discrete units as training targets and successfully apply S2UT to unwritten languages. \citet{translatotron2, inaguma-etal-2023-unity} propose two-pass S2ST models to improve the translation quality, which generate target text and target speech successively. ~\citet{zhang-etal-2024-streamspeech} introduces a unified framework for simultaneous speech recognition, translation and synthesis. To improve the decoding efficiency, \citet{inaguma2021non, xu-etal-2023-ctc} propose performing the S2TT task using non-autoregressive models. \citet{huang2023chch} first proposes NAR S2UT, achieving higher decoding speed but lower translation quality. \citet{wu-2023-duplex, zhu-etal-2023-diffs2ut} adopt diffusion models to iteratively generate discrete units. \citet{fang-etal-2023-daspeech} proposes the first two-pass NAR S2ST model, which achieves both good translation quality and fast decoding speed, but requires target text for training. \citet{fang-etal-2024-can} introduces a composite S2ST model which can combine existing speech-to-text translation and TTS models. \citet{ma-etal-2024-a} proposes a CTC-based NAR model for simultaneous S2ST. Compared with previous models, \name achieves better translation quality and faster decoding speed, and does not requires any text supervision during training.

\section{Conclusion}
In this paper, we propose \name, an NAR textless S2ST model based on CTC. By incorporating several training techniques including encoder pretraining, knowledge distillation, glancing training, and non-monotonic latent alignment, \name achieves comparable translation quality to the autoregressive S2UT model, while achieving 26.81$\times$ decoding speedup. Further ablation studies validate the effectiveness of each training technique.

\section*{Limitations}
Although \name achieves good performance in translation quality and decoding speed, it still has some limitations: (1) Training \name requires pretraining an autoregressive S2UT model, as its success relies on S2UT for encoder pretraining and knowledge distillation. This makes the training process more complex compared to S2UT; (2) Due to the limitations of existing publicly available datasets, we did not conduct experiments on real-world unwritten languages but instead simulated textless scenarios on language pairs such as Fr$\rightarrow$En, En$\rightarrow$Fr, and En$\rightarrow$Es.

\section*{Acknowledgement}
We thank all the anonymous reviewers for their insightful and valuable comments. This paper is supported by National Natural Science Foundation of China (Grant No.62376260).

\bibliography{custom}

\end{document}